\documentclass[conference]{IEEEtran}
\IEEEoverridecommandlockouts
\usepackage{cite}
\usepackage{amsmath,amssymb,amsfonts}
\usepackage{algorithmic}
\usepackage{graphicx}
\usepackage{textcomp}
\usepackage{xcolor}
\usepackage{tabularx}
\usepackage{marvosym}
\usepackage{caption} 
\usepackage{threeparttable}
\usepackage{textcomp}
\usepackage{diagbox}
\usepackage{booktabs}
\usepackage{multirow}

\def\BibTeX{{\rm B\kern-.05em{\sc i\kern-.025em b}\kern-.08em
    T\kern-.1667em\lower.7ex\hbox{E}\kern-.125emX}}
\begin{document}

\title{Composited-Nested-Learning with Data Augmentation for Nested Named Entity Recognition\\
}

\author{\IEEEauthorblockN{Xingming Liao}
\IEEEauthorblockA{\textit{School of Computers Science} \\
\textit{Guangdong University of Technology}\\
Guangzhou, China \\
2112205112@mail2.gdut.edu.cn}
\and
\IEEEauthorblockN{Nankai Lin}
\IEEEauthorblockA{\textit{School of Computers Science} \\
\textit{Guangdong University of Technology}\\
Guangzhou, China \\
neakail@outlook.com}
\and
\IEEEauthorblockN{Haowen Li}
\IEEEauthorblockA{\textit{School of Computers Science} \\
\textit{Guangdong University of Technology}\\
Guangzhou, China \\
2112305175@mail2.gdut.edu.cn}
\and
\IEEEauthorblockN{Lianglun Cheng}
\IEEEauthorblockA{\textit{School of Computers Science} \\
\textit{Guangdong University of Technology}\\
Guangzhou, China \\
0000-0002-8213-041X}
\and
\IEEEauthorblockN{Zhuowei Wang}
\IEEEauthorblockA{\textit{School of Computers Science} \\
\textit{Guangdong University of Technology}\\
Guangzhou, China \\
0000-0001-6479-5154}
\and
\IEEEauthorblockN{Chong Chen\textsuperscript{\Letter}}
\IEEEauthorblockA{\textit{School of Computers Science} \\
\textit{Guangdong University of Technology}\\
Guangzhou, China \\
0000-0003-2800-4647}
\thanks{\textsuperscript{\Letter} Chong Chen is the corresponding author.}
}

\maketitle

\begin{abstract}
Nested Named Entity Recognition (NNER) focuses on addressing overlapped entity recognition. Compared to Flat Named Entity Recognition (FNER), annotated resources are scarce in the corpus for NNER. Data augmentation is an effective approach to address the insufficient annotated corpus. However, there is a significant lack of exploration in data augmentation methods for NNER. Due to the presence of nested entities in NNER, existing data augmentation methods cannot be directly applied to NNER tasks. Therefore, in this work, we focus on data augmentation for NNER and resort to more expressive structures, Composited-Nested-Label Classification (CNLC) in which constituents are combined by nested-word and nested-label, to model nested entities. The dataset is augmented using the Composited-Nested-Learning (CNL). In addition, we propose the Confidence Filtering Mechanism (CFM) for a more efficient selection of generated data. Experimental results demonstrate that this approach results in improvements in ACE2004 and ACE2005 and alleviates the impact of sample imbalance.
\end{abstract}


\section{Introduction}

Named entity recognition (NER) is a fundamental task in information extraction that aims to identify text spans corresponding to specific types, such as person, organization, and locations. It plays an essential role in various downstream tasks and applications. Traditional approaches for NER rely on sequence labeling. In this type of method, each entity is treated as a separate and non-overlapping unit, and the goal is to assign entity labels to individual words or spans in the text \cite{luo2012features}. Nested Named Entity Recognition (NNER) offers greater flexibility compared to Flat Named Entity Recognition (FNER), enabling more fine-grained semantic representations and broader applications. However, a single word can have multiple labels, using these methods cannot effectively address NNER issues. To fill this gap, Tan et al. \cite{tan2020boundary} pointed out the importance of boundaries and addressed NNER using a boundary detection task. Fu et al. \cite{fu2021nested} employed a span-based constituency parser to address NNER. They treated the annotated entity spans as a partially observed constituency tree and incorporated latent spans during training.

\begin{table}
    \centering
    \caption{Label correlation analysis}
    \label{Relation}
    \renewcommand{\arraystretch}{1.3} 
    \setlength{\tabcolsep}{6pt} 
    
    \begin{threeparttable}
        \begin{tabular}{l|ccccccc}
        \hline 
        \diagbox[width=6.3em] {Inside}{Outside} & PER  & ORG  & GPE & FAC & WEA & LOC & VEH \\ 
        \hline
            PER           & 3018 & 1047 & 927 & 170 & 13  & 106 & 42  \\
            ORG           & 843  & 696  & 394 & 25  & 0   & 35  & 11  \\
            GPE           & 527  & 230  & 939 & 69  & 12  & 158 & 14  \\
            FAC           & 25   & 35   & 153 & 108 & 0   & 49  & 0   \\
            WEA           & 2    & 3    & 0   & 16  & 59  & 2   & 0   \\
            LOC           & 22   & 33   & 131 & 0   & 4   & 157 & 0   \\
            VEH           & 60   & 53   & 51  & 0   & 9   & 8   & 48  \\
            \hline
        \end{tabular}
        \begin{tablenotes}[flushleft]
            \footnotesize
            \item $Outside$ means containing nested labels, $Inside$ means nested labels
        \end{tablenotes}
    \end{threeparttable}
\end{table}

\begin{figure*}
\centering
\includegraphics[width=\textwidth]{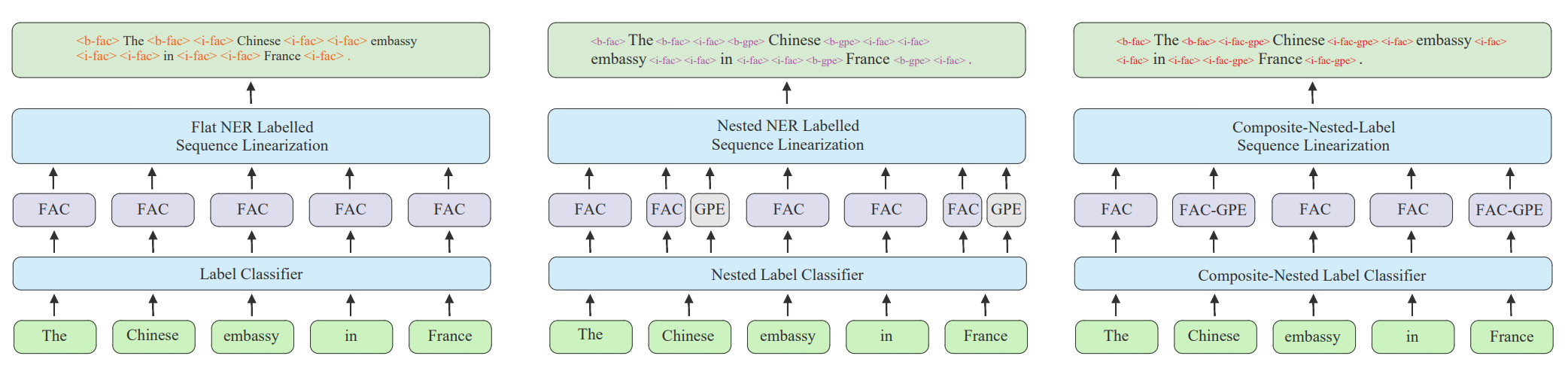}
\caption{Three different patterns for recognizing label sequence templates, with each distinguished by different colors for their respective label formats. The left part represents the Flat-NER classification template, the middle part represents the current  NNER classification template, and the right part represents the CNLC classification template we propose.} \label{FigCNLC}
\end{figure*}

However, the limited availability of annotated resources for nested entities constrains the performance of NNER. While some research efforts have focused on enhancing NER models through data augmentation. Ding et al. \cite{ding2020daga} introduced a novel data augmentation technique for sequence labeling tasks which can be an effective solution for low-resource NER. Using weakly labeled data augmentation method that can extract named entities from social media texts more effectively \cite{kim2022weakly}. To alleviate token-label misalignment issues, Zhou et al. \cite{zhou2021melm} proposed MELM as a new data augmentation framework for low-resource NER. But so far as I know, the use of data augmentation in NNER has been very limited, and existing data augmentation techniques cannot be directly applied to NNER tasks. 

In FNER, labels are assigned to tokens within a span, but it overlooks the possibility that one token may correspond to multiple labels, as shown in Fig. \ref{FigCNLC} (left). Labels containing nested labels are initially defined as $Outside$, with the nested labels being regarded as $Inside$. For example, consider the input $``The$ $Chinese$ $embassy$ $in $ $France$$"$. In this context, $``The$ $Chinese$ $embassy"$ is identified as the type of ``FAC”, while $``Chinese"$ is also assigned with the ``GPE" label. Consequently, in this case, the label ``FAC" is designated as $Outside$, and the label ``GPE" is categorized as $Inside$. As shown in Table \ref{Relation}, it can be observed that the ``PER" label is highly related to the ``PER" label and ``ORG" label. Simultaneously, there are cases where some labels have little or no correlation, so we can deduce that there is a certain level of correlation between labels. To fill this gap, we introduce a Composite-Nested-Label Classifier (CNLC) mechanism, which can simultaneously consider different labels for one token. Subsequently, the processed data is used for data augmentation through the Composited-Nested-Learning (CNL) module. As the augmented data exhibits the self-reinforcement effect \cite{yan2023understanding} not all augmented data positively impact the model. Therefore, we design a Confidence Filtering Mechanism (CFM) to select augmented data, and experiments indicate that this approach can provide more valuable data for NNER. To summarize, our main contributions are as follows:

\begin{itemize}
\item We composite nested tokens and nested labels in NNER for solving NNER problems through data augmentation methods.
\item For the data selection, our proposed CFM can select samples with higher confidence to improve the quality of the augmented data.
\item Through the framework, we can enhance existing models' performance, and it can alleviate the problem of sample imbalance. 
\item We open-source our augmented dataset, which could provide a sliver dataset for other researchers. 
\end{itemize}

\section{Related Works}

\subsection{Data augmentation}\label{AA}
In the field of Natural Language Processing (NLP), complex NER remains a relatively underexplored task. Data augmentation involves expanding the training dataset by applying transformations without changing the labels of the training instances. Using data augmentation methods effectively improves model generalization. Current research on data augmentation primarily addresses data scarcity in low-resource NLP, focusing on word-level modifications that prove beneficial for classification tasks, including simple synonym replacement strategies \cite{wei2019eda}, Masked Language Modeling (MLM) using PLMs \cite{kumar2020data} or auto-regressive PLMs \cite{kumar2020data}. One of the initial studies that delved into effective data augmentation for NER involved replacing named entities (NEs) with NEs of the same type or substituting tokens within the sentence with synonyms retrieved from WordNet \cite{dai2020analysis}. Subsequently, numerous neural learning systems were introduced, either altering the training objective of MLM using PLMs \cite{zhou2021melm} or mBART \cite{ghosh2023aclm}, to generate entirely new sentences from scratch. 

However, all these methods were designed for low-resource FNER, but for emerging and nested label complex entities, data augmentation in these domains remains underexplored.

\subsection{Nested NER}\label{BB}
Novel methods for NNER have recently been proposed and can be categorized into three main types: Sequence-to-sequence (Seq2Seq) methods, setting prediction methods \cite{li2023scoring}, and Span-based methods, respectively. Yan et al. \cite{yan2021unified} formulated NER as an entity span sequence generation problem, using a BART-based Seq2Seq model \cite{lewis2019bart} with a pointer network to solve most NER problems in a unified framework. Tan et al. \cite{tan2021sequence} proposed a non-autoregressive decoder for predicting entity sets, formulating NER as an entity set prediction task. Yuan et al. \cite{yuan2021fusing} focused on span-based methods and introduced three imitating mechanisms to handle heterogeneous factors in span-based methods.

Previous research on NNER has mainly studied the relationships between tokens, labels, and boundaries. The current data augmentation techniques are not sufficient to address the challenges of NNER. Therefore, we focus on labels of nested words within each span, combining different labels present within one word to address the limitation of data augmentation in NNER. Additionally, a data filtering mechanism is introduced to ensure that the generated samples are more readable and fluent.

\begin{figure*}
\centering

\includegraphics[width=\textwidth]{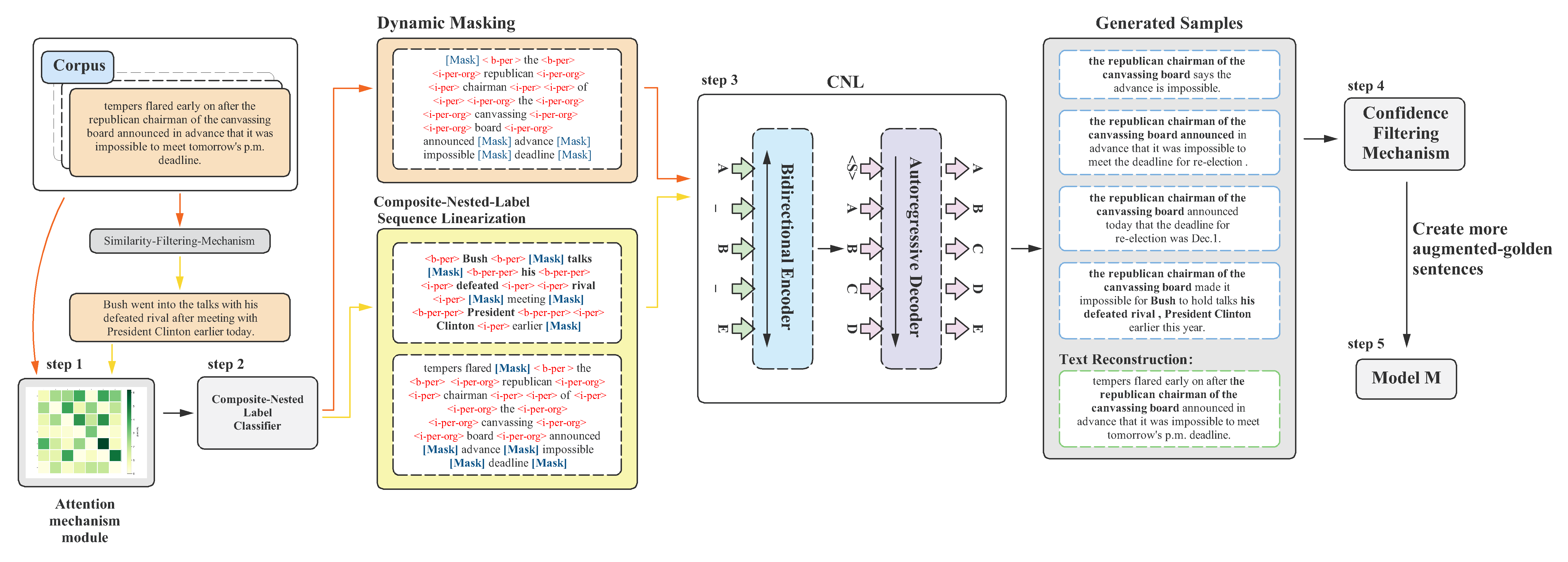}
\caption{Model architecture of CNL: Model CNL is divided into five steps, serving as the input to the model during fine-tuning and generation. Step 1: Similar sentences are obtained from the corpus using a similarity filtering mechanism. Then, the important keywords related to NEs are extracted using attention maps obtained from the fine-tuned RoBERTa model. Step 2: After adding label tokens before and after each entity in the sentences using CNLC, the sentences are divided into two parts. The original sentence template undergoes further masking, with a small portion of keywords dynamically masked. The other part is obtained by merging the sentence with similar sentences using a FUSION mechanism to create a template. Step 3: The CNL model is used to generate augmented data. Step 4: The generated samples are further filtered through the CFM to obtain high-confidence sentences, which are then concatenated with the golden data. Step 5: The obtained final data is used as the input for model $M$.}\label{FigCNL} 
\end{figure*}

\section{Method}
In this section, we first introduce the label classifier about the CNLC in \ref{3.1} and then describe our method for data augmentation in \ref{3.2}. The overview of the CNL augmentation is shown in Fig.  \ref{FigCNL}. The augmentation data requires selecting the CFM, and the selecting mechanism will be introduced in \ref{3.3}.

\subsection{Composite-Nested-Label Classification Template}\label{3.1}

Previous NNER studies have primarily focused on spans and tokens as they are distinctive markers with rich semantics, with the primary goal of identifying whether a span constitutes an entity. Many methods have been explored how accurately models traverse or enumerate all spans. As shown in Fig. \ref{FigCNLC} (middle), given an input sequence $``The$ $Chinese$ $embassy$ $in $ $France$$."$, enumerates each span to populate $<$label$>$ of the template. In the data augmentation field, the approach can not address the issue of a single token having multiple labels in NNER. However, we've discovered that for nested words, the label of the word itself is a crucial determinant. From our Table \ref{Relation}, It’s evident that there is a correlation among labels. Therefore, We introduce CNLC and use it to construct a new template. Combining tokens and labels, allowing the model to learn labels for span perception and considering different labels' contributions to tokens, as shown in Fig. \ref{FigCNLC} (right). With this template, we can obtain the nested labels with a span by traversing it only once and can effectively address the problem of NNER in data augmentation.

\subsection{Composited-Nested-Learning Model}\label{3.2}
For every sentence in our training dataset, we initially identify a set of none-Named Entity (n-NE) tokens that receive the highest attention from the Named Entities (NEs) within that sentence. These selected tokens are termed as keywords. In our research, an n-NE token is defined as a keyword when it holds the most contextual significance for the NEs within the sentence. The contextual dependency is assessed by utilizing the attention scores extracted from attention maps derived from a RoBERTa model that we fine-tuned with the golden dataset. Our primary objective is to identify the top k\% of n-NE tokens, referred to as keywords. To ensure robustness, we make certain that no more than 10\% of their combined attention is directed towards a single token \cite{clark2019does}. Additionally, stop words, punctuation, and other NEs are excluded from the top k\% of n-NE tokens to derive the ultimate set of keywords. Once the top k\% of n-NE tokens in the sentence are designated as keywords, we end up with the K n-NE keywords and E entity tokens. To construct the template, we replace each n-NE token that is not part of the K keywords with a mask token and then eliminate consecutive mask tokens.

Inspired by Zhou et al. \cite{zhou2021melm}, we perform CNLC sequence linearization on the CNL template proposed in Section \ref{3.1}, and subsequently incorporate label information into the fine-tuning and augmentation generation process. Similar to the approach \cite{zhou2021melm}, Label tokens are inserted both before and after each entity token, considering them as part of the regular context within the sentence. Furthermore, these label tokens positioned both before and after each NE offer boundary supervision, particularly for NEs that span multiple tokens.

After CNLC sequence linearization, This process is initiated by sampling a dynamic masking rate $\epsilon$ from a Gaussian distribution: 
\begin{equation}
 f(x) = \frac{1}{\sigma\sqrt{2\pi}} e^{-(x-\mu)^2/(2\sigma^2)}
\end{equation} 

where the variance $\sigma$ is set to $\frac{1}{K}$, and $x$ is the keyword index list of the CNLC processed template. 

Subsequently, we randomly select tokens from the set of $K$ keywords in the sentence based on the masking rate $\epsilon$, replace them with mask tokens, and then eliminate consecutive mask tokens. The aforementioned process serves two primary purposes: 1) It allows for the creation of different templates, leading to the generation of a more diverse set of sentences and enhancing sentence diversity. 2) The varied combinations of keywords and entities activate the model's generation capacity.

As previously explained, the CNL model is fine-tuned on text with missing information, and it learns to recover the original text from the templates. This is employed as our fine-tuning objective, and templates are created that deviate from existing pre-training objectives through our strategy.

The sentences involved in the aforementioned process are referred to as sentence A. To retrieve sentence B similar to A, we traverse all sentences in the dataset and identify the top-n sentences with the highest similarity to A. To calculate the semantic similarity for each sentence in the training set, embeddings for each sentence are initially extracted using a multi-lingual Sentence-BERT \cite{reimers2019sentence}. Then compute the semantic similarity using the following method: 
\begin{equation}
\text{sim}(S_a, S_b) = \frac{{S_a \cdot S_b}}{{\|S_a\| \, \|S_b\|}}
\end{equation}

where $sim(\cdot)$ is the cosine similarity between two embeddings and \(S_a\), \(S_b\) respectively represent the aforementioned Sentence $A$, $B$. In addition, $a$, $b$ $\in$ $N$ where $a$ $\neq$ $b$ and $N$ is the size of the training set.

Upon obtaining the similar sentence $B$, we apply the aforementioned missing information template operation. After acquiring templates for both sentences $A$ and $B$, we propose the use of FUSION, a novel template fusion algorithm that combines sentence $A$ and sentence $B$, resulting in the creation of a new context-enriched sentence named $C$. Similarly, $C$ is input into the CNL model, allowing the sentence to be regenerated by the model.

\subsection{Confidence Filtering Mechanism}\label{3.3}
\begin{figure}
\centering
\includegraphics[width=0.75\linewidth]{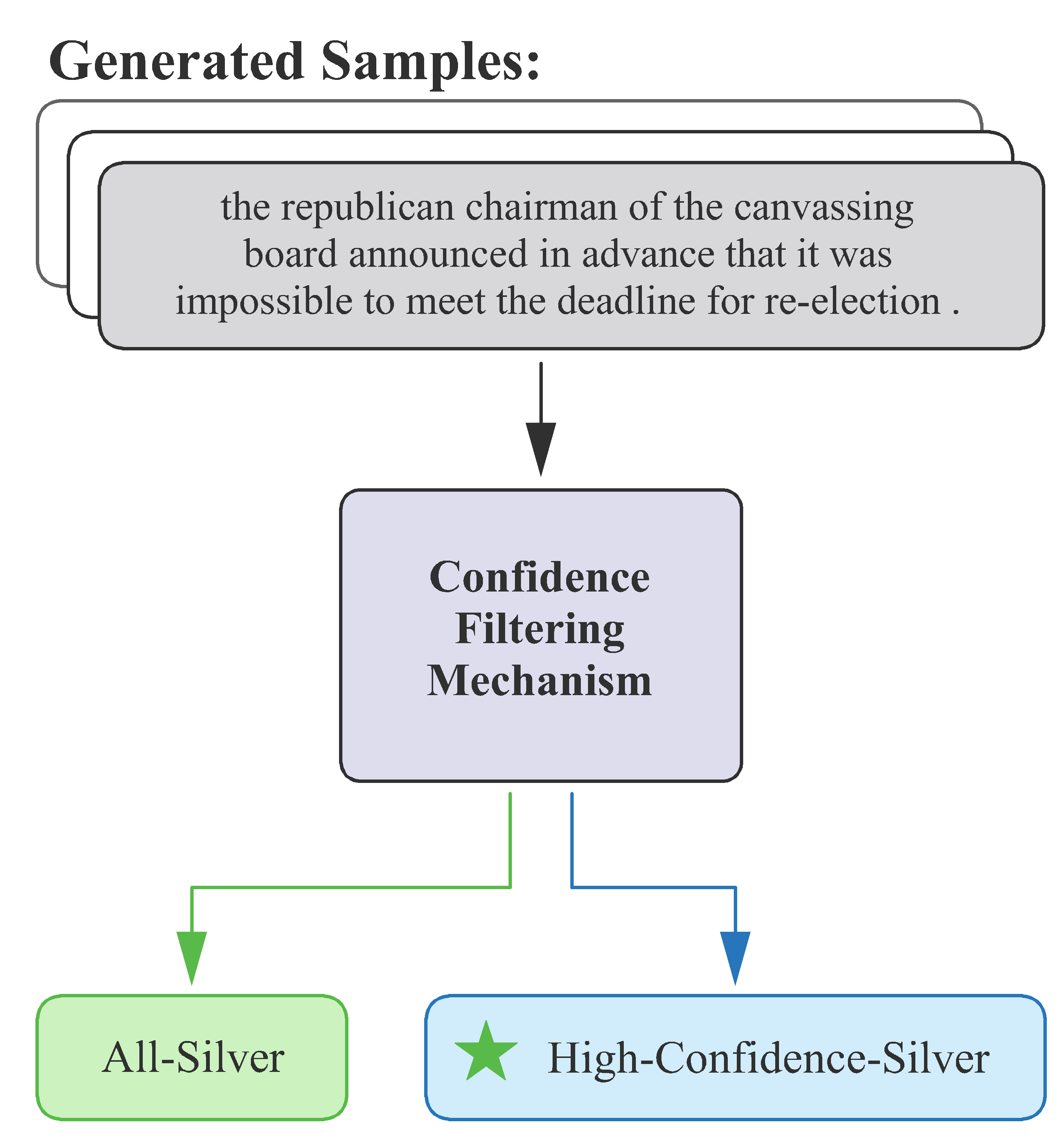}
\caption{After acquiring data-augmented samples, the samples are subsequently filtered using the CFM. Within the sample filtering process, sentences with low PLLs are excluded, and high-confidence sentences are retained as our final silver dataset.} \label{CFM}
\end{figure}

As a post-processing step, following the operation in Section \ref{3.2}, an augmented dataset is obtained. Initially, we calculate the accuracy of the model's predicted-label compared to the golden-label of the original sentences. This results in two subsets: “silver” samples and “none-silver” samples. It is important to note that the “silver” category may contain sentences where only the labels match, but the sentences have low pseudo-log-likelihood scores (PLLs) \cite{salazar2019masked}. Therefore, we propose CFM to effectively alleviate the issues mentioned above. By inputting the generated samples into a module and filtering them based on PPLs, we can construct a ``high-confidence silver'' dataset as shown in Fig. \ref{CFM}. When the ``high-confidence silver'' samples are concatenated with the original training set samples, we obtain an enriched NNER dataset. We can input the augmented data into Model $M$ as part of its input.

\begin{table}[]
\caption{Statistics and silver for ACE2004 and ACE2005}
\label{dataset}
\setlength{\tabcolsep}{8pt}
\renewcommand{\arraystretch}{1.15}
\begin{threeparttable}
\begin{tabular}{@{}ccccccccl@{}}
\toprule
     & \multicolumn{3}{c}{ACE2004} & \multicolumn{1}{l}{} & \multicolumn{3}{c}{ACE2005} &    \\ \cmidrule(l){2-4} \cmidrule(l){6-8} 
     & Train    & Dev     & Test   &  & Train    & Dev     & Test   &  \\   \cmidrule(r){1-9}
\#S  & 6198     & 742     & 809    &  & 7285     & 968     & 1058   &  \\
\#NS & 2718     & 294     & 388    &  & 2797     & 352     & 339    &  \\
\#E  & 22204    & 2514    & 3035   &  & 24827    & 3234    & 3041   &  \\
\#NE & 10159    & 1092    & 1417   &  & 10039    & 1200    & 1186   &  \\    \cmidrule(r){1-9}
*S   & 5910     & -     & -    &  & 5404     & -     & -   &  \\   
*NS  & 1062     & -     & -    &  & 983      & -     & -    &  \\
*E   & 14693    & -    & -   &  & 13778    & -    & -   &  \\
*NE  & 1676     & -    & -   &  & 1416     & -    & -   &  \\ \bottomrule
\end{tabular}
\begin{tablenotes}[flushleft]
\footnotesize
\item \#represents Statistics, and * represents Silver. We report the number of sentences(\#S,*S), the number of sentences containing nested entities(\#NS,*NS), the number of entities(\#E,*E), and the number of nested entities(\#NE,*NE) on the two datasets.
\end{tablenotes}
\end{threeparttable}
\end{table}

\begin{table*}[]
\caption{Results in the standard nested NER setting.}
\label{standard}
\setlength{\tabcolsep}{12pt}
\renewcommand{\arraystretch}{1.20}
\begin{threeparttable}
\begin{tabular}{@{}lcccccccl@{}}

\toprule
\multirow{2}{*}{\large{Model}}                          & \multicolumn{3}{c}{ACE2004} & \multicolumn{1}{l}{} & \multicolumn{3}{c}{ACE2005} &  \\ \cmidrule(l){2-4} \cmidrule(l){6-8} 
                                                & Pr.     & Rec.    & F1      &                      & Pr.     & Rec.    & F1      &  \\ \cmidrule(r){1-9}
Biaffine \cite{yu2020named}                    & 87.30    & 86.00      & 86.70    &                      & 85.20    & 85.60    & 85.40    &  \\
BARTNER \cite{yan2021unified}                      & 87.27   & 86.41   & 86.84   &                      & 83.16   & 86.38   & 84.74   &  \\
UIE \cite{lu2022unified}                           & -       & -       & 86.89   &                      & -       & -       & 85.78   &  \\
BuParser \cite{yang2021bottom}                               & 86.60    & 87.28   & 86.94   &                      & 84.61   & 86.43   & 85.53   &  \\
ERPG \cite{zhu2023erpg}                         & -       & -       & 86.99   &                      & -       & -       & 86.38   &  \\
PromptNER* \cite{shen2023promptner} & \underline{86.91}   & \underline{87.67}   & \underline{87.29}   &                      & \underline{85.22}   & \underline{88.03}   & \underline{86.60}    &  \\
CNL                                             & \textbf{87.12 (+0.21)}   & \textbf{88.27 (+0.60)}   & \textbf{87.69 (+0.40)}   &                      & \textbf{85.93 (+0.71)}   & \textbf{88.60 (+0.57)}    & \textbf{87.25 (+0.65)}   &  \\ \bottomrule

\end{tabular}
\begin{tablenotes}[flushleft]
\footnotesize
\item Bold and underline indicate the best and the second-best scores
\item \textsuperscript{*}Baseline result given by our implementation. Other results are derived from related papers
\end{tablenotes}
\end{threeparttable}
\end{table*}

\section{Experiments and Results}
\subsection{Datasets}\label{AA}

We conducted experiments on two datasets: ACE2004 and ACE2005. During the data augmentation phase, label filtering was initially performed on the ACE2004 and ACE2005 datasets. Because overly complex situations were not considered, sentences in which a single token has more than three nested labels were excluded. After data augmentation, the augmented corpus was concatenated with the original training data, resulting in an aug-golden dataset. Following previous work, we measured the results using span-level precision, recall, and F1 scores. Models were selected based on the performance of development sets. We followed Yu et al. \cite{yu2020named} to train the model on the concatenation of the train and dev sets. Table \ref{dataset} provides relevant information about the dataset.

\subsection{Experimental Setup}\label{BB}
\textbf{CNL. } XLM-RoBERTa-large with an additional linear head was employed in our attention selection CNL model. We approach the task as token-level classification, implementing the BIO tagging scheme. Our model optimization is carried out using the Adam optimizer, with a fixed learning rate of 1e$^-$$^2$, and training is performed with a batch size of 8. The attention selection CNL model is trained for 100 epochs, with an attention mask rate of 0.3. The model that exhibits the best performance on the development set is selected for testing.

We use mBart-50-large equipped with a condition generation head to enhance the performance of CNL. The fine-tuning process for CNL spans 10 epochs, facilitated by the Adam optimizer with a learning rate of 1e$^-$$^5$. A batch size of 16 is used in dataset ACE2004, and a batch size of 8 is used in dataset ACE2005. In the FUSION module, the attention mask rate is set to 0.3.

\textbf{PromptNER. } PromptNER has been selected as our model $M$, and this model serves as our baseline. Our results are compared based on our reimplementation of PromptNER. As a result of the addition of more samples, the decision has been made to extend the training period for better refining the model. The original epoch value has been adjusted from 50 to 60 while keeping all other parameters unchanged. Similarly, we have chosen recent competitive models as our baseline, including parsing-based \cite{yu2020named}, generation-based \cite{yan2021unified} \cite{lu2022unified}, span-based \cite{yang2021bottom} \cite{yuan2021fusing}, Prompt Guidance \cite{zhu2023erpg} \cite{shen2023promptner}. These approaches utilize different pre-trained language models as the encoder. Therefore, in our experimental results, performance is presented using BERT-large as the benchmark.

\textbf{Result. }  Table \ref{standard} illustrates the performance of data augmentation through CNL, the performance of the current state-of-the-art (SOTA) models in the NNER field can be further improved. The silver dataset can be extended to other models.

\subsection{Parametric Search}\label{CC}

\begin{figure}
\centering
\includegraphics[width=0.5\textwidth]{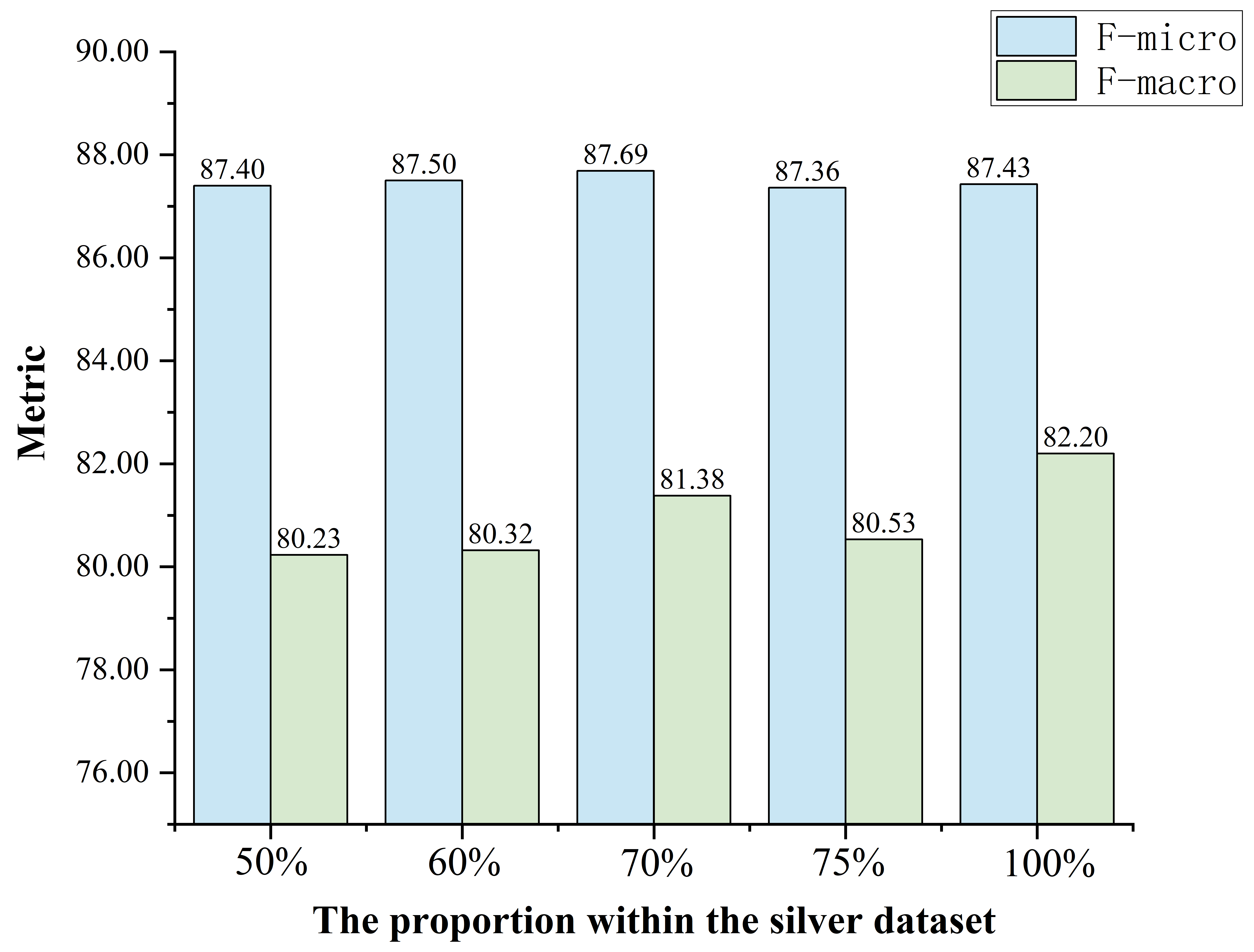}
\caption{A parameter search was conducted for the silver dataset generated for ACE2004. The Rate represents the proportion of silver selected, and we tested it using model $M$, obtaining F-micro and F-macro scores at different proportions.}\label{para} 
\end{figure}

After enhancing the data, the presence of sentences in the new silver dataset containing repeated occurrences of simple words or phrases was observed. Although the predicted labels matched the golden labels, these sentences lacked coherence and had low reference values. To address these low-confidence samples, the CFM was introduced to assign confidence scores to the samples. The higher the confidence score, the more complete the sentence. To select the best samples, we varied the proportion of silver samples (50\%, 60\%, 70\%, 75\%). As shown in Fig. \ref{para}, it can be noted that selecting 70\% of the silver samples in ACE2004 yielded the best results. Using the same approach, we achieved the optimal outcome by choosing 35\% of the silver samples in ACE2005.


\subsection{Sample Imbalance Inquiry}\label{DD}

Following data augmentation with the CNL model, it is observable from Table \ref{Relation} that the distribution of the dataset tends to exhibit bias towards the first few classes, leading to an excessive leaning of the model towards the majority class. Through the data augmentation methods presented in Table \ref{equilibrium},  We appropriately expanded some underrepresented labels, which alleviated the problem of sample imbalance to a certain extent and enhanced the balance of the model among different categories. Similarly, alleviating sample imbalance can help the model generalize better to diverse data from various categories, thereby improving the model's practicality and robustness.

\begin{table}[]
\caption{Results in sample imbalance inquiry.}
\setlength{\tabcolsep}{9pt}
\renewcommand{\arraystretch}{1.2}
\label{equilibrium}
\begin{threeparttable}
\begin{tabular}{lcccc}
\toprule
\multicolumn{1}{c}{} & ACE2004 & ACE2004* & ACE2005 & ACE2005* \\ \midrule
WEA                  & 75.29   & 71.60    & 78.85   & 79.21    \\
PER                  & 91.04   & 91.50    & 90.01   & 90.29    \\
FAC                  & 71.53   & 67.67    & 77.66   & 77.70    \\
GPE                  & 89.57   & 89.64    & 85.96   & 88.50    \\
ORG                  & 81.85   & 82.58    & 82.66   & 82.50    \\
VEH                  & 91.67   & 100.00   & 74.49   & 75.90    \\
LOC                  & 64.47   & 66.67    & 70.09   & 74.77    \\
Fmacro               & 80.77   & \textbf{81.38}    & 79.96   & \textbf{81.27}    \\ \bottomrule
\end{tabular}
\begin{tablenotes}[flushleft]
\footnotesize
\item ACE2004 and ACE2005 for model PromptNER* results given by our implementation. ACE2004* and ACE2005* for CNL.
\end{tablenotes}
\end{threeparttable}
\end{table}

\section{Conclusion}
This work introduces a CNL model for NNER and its application in data augmentation. We demonstrate the existence of correlations between labels, and based on these label correlations, we propose CNLC templates to address the challenge of applying NNER in data augmentation. For the enhancement of generated silver samples, we introduce the CFM, effectively helping us filter sentences with high confidence. Our approach focuses on how to augment NNER samples, making them applicable to other models and providing more valuable samples. Through these samples, it can improve the model's performance and alleviate sample imbalance. In future work, we will endeavor to apply our method to few-shot NNER and explore methods for addressing more complex nested entity labels.

\section{Acknowledgment}
Our work is supported by multiple funds in China, including Guangzhou Science and Technology Planning Project (2023B01J0001), the Key Program of NSFC-Guangdong Joint Funds (U2001201, U1801263), Industrial core and key technology plan of Zhuhai City (ZH22044702190034-HJL). Our work is also supported by Guangdong Provincial Key Laboratory of Cyber-Physical System (2020B1212060-069).

\bibliographystyle{IEEEtran}
\bibliography{refs}

\end{document}